\documentclass{article}
\usepackage{spconf,amsmath,graphicx}
\usepackage{amssymb}

\title{TRACKING OF ENRICHED DIALOG STATES FOR FLEXIBLE CONVERSATIONAL INFORMATION ACCESS}
\name{Yinpei Dai, Zhijian Ou$$, Dawei Ren, Pengfei Yu \thanks{This work is supported by NSFC grant 61473168.}}
\address{Speech Processing and Machine Intelligence (SPMI) Lab, Tsinghua University, Beijing, China\\
dyp16@mails.tsinghua.edu.cn, ozj@tsinghua.edu.cn}
\begin{document}
\ninept
\maketitle
\begin{abstract}
Dialog state tracking (DST) is a crucial component in a task-oriented dialog system for conversational information access.
A common practice in current dialog systems is to define the dialog state by a set of slot-value pairs.	
Such representation of dialog states and the slot-filling based DST have been widely employed, but suffer from three drawbacks.
(1) The dialog state can contain only a single value for a slot, and (2) can contain only users' affirmative preference over the values for a slot.
(3) Current task-based dialog systems mainly focus on the searching task, while the enquiring task is also very common in practice. 
The above observations motivate us to enrich current representation of dialog states and collect a brand new dialog dataset about movies, based upon which we build a new DST, called enriched DST (EDST), for flexible  movie information access.
The EDST supports the searching task, the enquiring task and their mixed task. 
We show that our new EDST method not only achieves good results on Iqiyi dataset, but also outperforms other state-of-the-art DST methods on the traditional dialog datasets, WOZ2.0 and DSTC2.

\end{abstract}
\begin{keywords}
Dialog state, dialog state tracking, recurrent neural network, dialog dataset
\end{keywords}
\section{Introduction}
\label{sec:intro}

Dialog state tracking (DST) is a crucial component in a task-oriented dialog system for conversational information access.
The dialog state summarizes the dialog history, including all previous user utterances and all system actions taken so far. It is passed to the system's dialog policy that decides which action to take.
In general, the dialog state consists of elements with human-interpretable meanings from the task ontology, which describes the scope of semantics the system can process.
The ontology is often specified by a collection of \textit{slots} and the \textit{values} that each slot can take. 
So a common practice in most dialog systems is to define the dialog state by a set of slot-value pairs.
For example, in a movie information access system, after the user utterance \textsl{``I want to see Nolan's thriller"}, the dialog state gets updated to consist of new slot-value pairs \textsl{``director=Nolan; genre=thriller"}. In this case, we often say the dialog state contains the value ``\textit{Nolan}" for slot ``\textit{director}", and the value ``\textit{thriller}" for slot ``\textit{genre}".
To accommodate uncertainty, most modern dialog state tracker maintains a probability distribution over dialog states, which is often called belief state.
In practice, the tracker updates a multinomial distribution over all possible values for each slot separately, which is thus often referred to as slot-filling.

\begin{figure}[htp]
\centering
\centerline{\includegraphics[width=8cm]{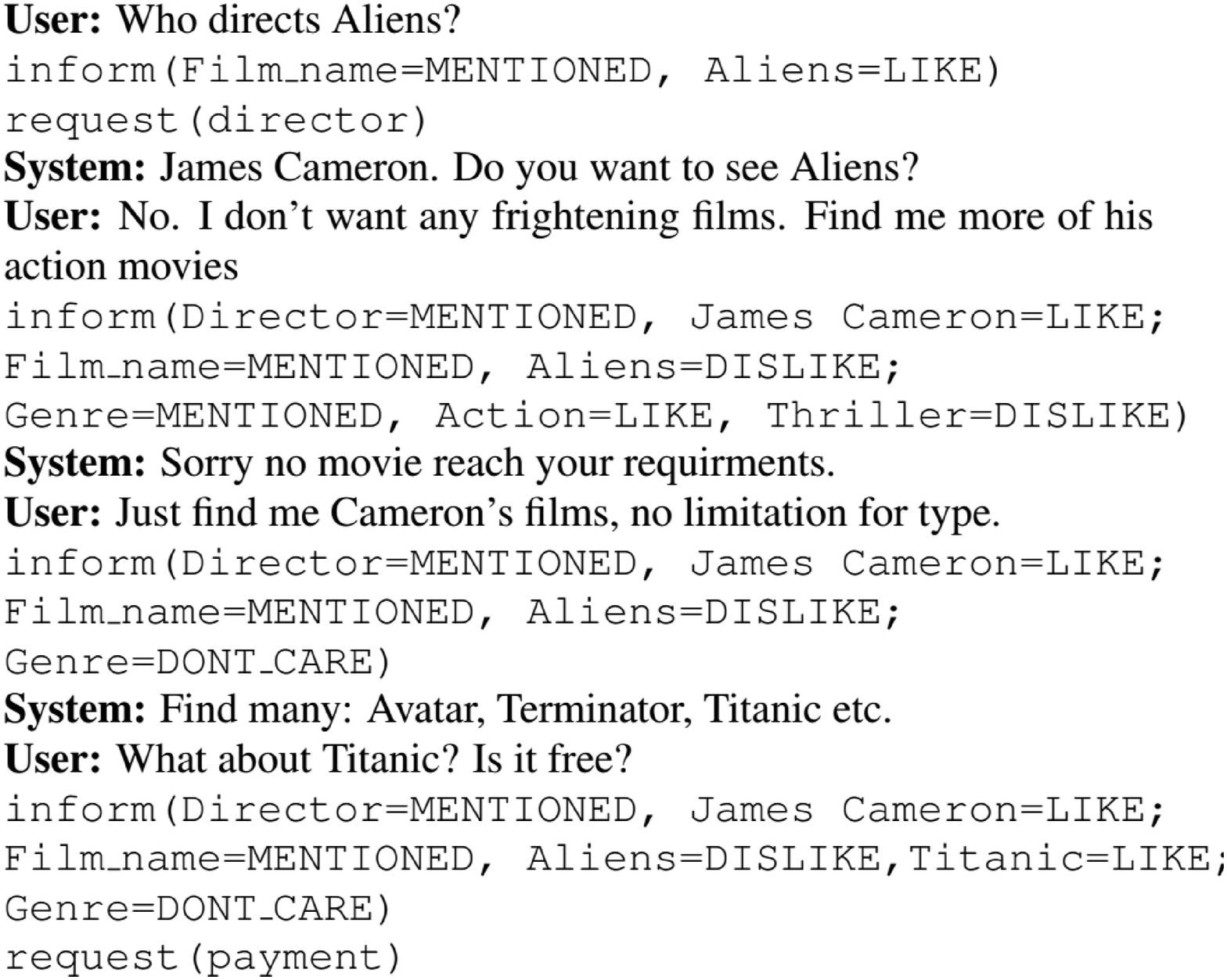}}
\caption{Example of the Iqiyi dialog dataset with annotated new dialog states. Slots and values with label \texttt{NOT\_MENTIONED} are not shown for simplicity. \texttt{inform} is related with informable slots, \texttt{request} is pertinent to requestable slots.}
\label{fig:dialog_example}
\vspace{-0.5cm}
\end{figure}

Such representation of dialog states and the slot-filling based DST have been widely employed, e.g. in the Dialogue State Tracking Challenge (DSTC) series \cite{the-dialog-state-tracking-challenge-series-a-review,Kim2017Fourth,Kim2017Fifth}. However, this standard practice suffers from three drawbacks.
First, in the set of slot-value pairs which defines the dialog state, at most one slot-value pair are allowed for an informable slot\footnote{Informable slots are slots that users can use to constrain the search, such as movie type.}, or say, the dialog state can  contain only a single value for a slot.
For example, user's sentence \textsl{``I want moderately cheap restaurant"} can only be labeled as either \textsl{``price range=moderate"} or \textsl{``price range=cheap"} but not both in the DSTC2 dataset; in fact, the user wants a cheap or moderate restaurant, and both slot-value pairs should be included in DST. So it is desirable to enrich current dialog state representation to contain multi-values.

Second, when the dialog state contains a value for a slot, the preference is by default to be affirmative, or say, the dialog state can contain only users' affirmative preference over the values for a slot. 
However, in real-world information access, negative preferences are often expressed by users.
Although in DSTC2, users' preference over values is partially supported by extra tags like "\textit{User Actions}", this requires extra efforts to label data and build DST. So it is desirable to enrich current dialog state representation to contain such preference over values.

Third, current popular task-oriented dialog systems and datasets mainly focus on the searching task \cite{wenN2N17,Dhingra2017Towards}.
The system helps the user to find certain entity by interactively asking for attributes which helps constrain the search (i.e. informable slots). Once found, the user can retrieve information by asking for requestable slots\footnote{Requestable slots are slots that users can ask a value for, such as movie length.}.
In practice, however, the enquiring task in which the user directly asks for the requestable slots of some movies is also very common.

The above observations motivate us to enrich current representation of dialog states and collect a brand new dialog dataset about movies, based upon which we can build a flexible dialog system for accessing movie information. The system supports the searching task, the enquiring task and their mixed task. 
To overcome the limited representation capability of slot-value pairs, we propose to use two sets of labels for slots and values separately, so that the dialog state can contain multi-values and express preference over values (like/dislike) for a slot.
Moreover, a novel DST, called enriched DST (EDST), is developed for tracking the enriched dialog states and supporting more flexible information access.
We show that the new EDST method not only achieves good results on Iqiyi dataset, but also outperforms other state-of-the-art DST methods on the traditional dialog datasets, WOZ2.0 \cite{Mrk2016Neural} and DSTC2 \cite{the-dialog-state-tracking-challenge-series-a-review}.

\section{IQIYI dialog DATASET}
\label{sec:dataset}

We aim to develop a text-in text-out dialog dataset that contains both searching tasks (e.g. \textsl{usr: ``I want to see Cameron's recent film". sys: ``Find Avatar"}) and enquiring tasks (e.g. \textsl{usr: ``What's Avatar's director". sys: ``Cameron"}). In this dataset, the user is able to change his/her goals at will, and proposes searching constraints freely. Since this is already a non-trival task, we do not take ASR into account. \textit{Movie} is chosen as our dialog domain. The movie database is extracted from Iqiyi movie website \footnote{http://www.iqiyi.com/dianying/}, which gives the name of our new dataset. A crowdsoucing website, similar to the Wizard-of-Oz (WOZ) website \cite{wenN2N17}, was built to collect the dialog data. All the dialog data are in Chinese. After careful cleaning of the raw data, we collected 800 dialogues in total \footnote{
http://oa.ee.tsinghua.edu.cn/ouzhijian/data/Iqiyi\_movie\_data.rar}.

\textbf{Ontology.} There are 7 informable slots in total
: \textit{Film\_name, Director, Actor, Genre, Country, Time, Payment.} All the informable slots can take multiple values, except \textit{Film\_name} can take only one value which the user would prefer, since it is more natural for the user to focus on just one favourite movie entity at one turn in the task-oriented system. 11 requestable slots are chosen in total, 7 from informable slots and 4 extra are \textit{Release\_date, Critic\_rating, Movie\_length, Introduction.}

\textbf{Enriched Dialog States.} We enrich the traditional dialog state representation, by employing two sets of labels for slots and values seperately. In this dataset,  we use  \texttt{DONT\_CARE, MENTIONED, NOT\_MENTIONED} as the labels for each informable slot since they can represent the general status of slots through conversations, and we choose the most common labels \texttt{LIKE, DISLIKE, NOT\_MENTIONED} for each value. Our new dialog state takes the form as slot-label pairs and value-label pairs. In this tagging scheme, the dialog state could include multiple values and user's polarity preference directly. Figure \ref{fig:dialog_example} illustrates an example piece of Iqiyi dialog dataset with annotated new dialog states.

\begin{table}[htp]
\centering
\begin{tabular}{|l|l|l|l|}
\hline
dataset & WOZ2.0  & DSTC2 & Iqiyi\\
\hline
domain & restaurant & restaurant & movie\\
\#words   & 784  & 1739 & 1527 \\
\#named entity & 113  & 113 & 147 \\
\#slots & 3  & 4 & 7 \\
\#values  & 212 & 212 & 599\\
\#synonyms & $\approx$ 60 &  $\ll$ 60  &  $\approx$ 200 \\
goal change & 11.2\% & 6.24\% & 13.1\%\\
\hline
\end{tabular}
\caption{Comparision among datasets. \#slots is the number of informable slots, \#values is the number of all the values in ontoloty, \#named entity is the number of restaurants or movies in the database. Goal change is the percentage of turns in which the user changes his/her mind.}
\label{table1}
\vspace{-0.5cm}
\end{table}

\textbf{Dialog Tasks.} Three types of task descriptions are constructed in the WOZ experiment as the guidence for collecting dialog data: (1) Seaching task: ask the user to find an unknown movie while knowing some attributes of it; (2) Equiring task: given a known movie, the user need to find out some attributes of it; (3) Mixed task: while knowing a movie's name or some attributes, the user needs to search more related unknown movies and their attributes. In the data collection, both users and wizards do not have to follow the task guidance strictly, and are encouraged to provide any kind of conversation as long as our new dialog state can handle. 	

\textbf{Comparison.} Two typical dialog datasets, WOZ2.0 \cite{Mrk2016Neural} and DSTC2 \cite{the-dialog-state-tracking-challenge-series-a-review}, are chosen for comparison since they are freely available, and the existing DST methods tested over them become baselines for studying our new DST model.
Compared to WOZ2.0 and DSCT2, the Iqiyi dialog dataset 
not only has more complex dialog states,
but also consists of more lexicon variations and goal changes, as shown in Table \ref{table1}. 
This makes the Iqiyi dialog dataset a more chanllenging testbed for searching and enquiring dialog tasks.

\section{DST MODEL}
\vspace{-0.4em}
\label{sec:model}

Enriched Dialog State Tracker (EDST) is a Jordan-type RNN customized for our Iqiyi dialog dataset. Its recurrent component consists of two sub-trackers, value-specific tracker (VST) and slot-specific tracker (SST), which are built for tracking value labels and slot labels separately. Since multiple values are allowed to be chosen,  all the values are predicted separately to avoid state space exploding . 

\subsection{Model Definition}
\label{sec:model_definition}
Let $s$ denote the slot entity (e.g. \texttt{Genre}), $v$ denote the value entity (e.g. \texttt{Thriller}), and $V^s$ denote the vocabulary of all value entities for slot $s$. $S$ denotes the vocabulary of all slot entities. Let $\eta_{v}$ denote the tracking variable corresponding to value $v$ at current turn, so $\eta_{v}$ can take \texttt{LIKE, DISLIKE, NOT\_MENTIONED}. $\eta^s$ denotes the set of variables $\{\eta_{v}: v \in V^s\}$, and $\eta$ represents the set $\{\eta^s: s\in S\}$. We have similar notations for slots. Let $\xi_s$ denote the tracking variable corresponding to slot $s$ at current turn, thus $\xi_s$ can take \texttt{DONT\_CARE, MENTIONED, NOT\_MENTIONED}. $\xi$ is defined as the set $\{\xi_s: s \in S\}$. 

Assuming that the dialog is Markovian, we denote the user uterrance at current turn by ${u}$, the system act and belief state at the last turn by ${a}$ and ${b}$ respectively. 
The belief state at current turn is defined by $p(\xi,\eta|u,a,b)$, the posterior distribution of $\xi,\eta$ given $u,a,b$.
The main purpose of our DST is to maintain this belief state at each turn, which is recursively updated as follows:
\footnote{We drop the condition $u,a,b$ and replace label \texttt{NOT\_MENTIONED} with \texttt{NOT\_MEN} in the following to simplify the notation.}
\begin{equation}
\setlength\abovedisplayskip{1pt}
\setlength\belowdisplayskip{1pt}
\begin{split}
p(\xi,\eta)
=&\prod_{s \in S}p(\xi_s,\eta^s)=
\prod_{s \in S}p(\xi_s|\eta^s)p(\eta^s)\\
=&\prod_{s\in S}\left(p(\xi_s|\eta^s)\prod_{v\in V^s}p(\eta_v)\right)
\end{split}
\end{equation}
It can be easily seen that the independence among variables underlying the above factorization is reasonable. 
Also note that in traditional DST, the belief state for slot $s$ is represented by a multinomial distribution over values $v\in V^s$. In contrast, in EDST, it is represented by $p(\xi_s,\eta^s)$, which is clearly more expressive and flexible.

In the following, we design two trackers, the value-specific tracker (VST) and slot-specific tracker (SST), for tracking $p(\eta_v)$ and $p(\xi_s|\eta^s)$ respectively.

\subsection{Value-Specific Tracker}
\label{sec:VSC}
In VST, for each slot $s$, we iterate over all possible values $v\in V^s$ to update $p(\eta_v)$, and finally we have $p(\eta^s)=\prod_{v\in V^s}p(\eta_v)$. 
So the basic operation is to update $p(\eta_v)$ for $v\in V^s$, which contains three main steps as described in the following. Its input consists of last system act $a$, last belief states $b$, current user utterace $u$, and the value $v$ over which we need to update the posterior distribution $p(\eta_v)$.

First, we convert the input to value-specific features.

1. From the last belief states ${b}$, a 3-dimensional value-specific belief vector is extracted by $f_1(v,\cdot)$ operation, which carries relevant information only to $v$:
\begin{equation}
\setlength\abovedisplayskip{1pt}
\setlength\belowdisplayskip{1pt}
\begin{split}
f_1(v,b)=&\left(p(\eta_{v}^{'}=\texttt{LIKE}),p(\eta_v^{'}=\texttt{DISLIKE}),\right.\\
& \left.p(\eta_{v}^{'}=\texttt{NOT\_MEN})\right)^T
\end{split}
\label{value-specific state vetor}
\end{equation}
where $\eta_v^{'}$ denotes the variable corresponding to $v$ at the last turn. $T$ denotes matrix transpose.
 
2. From the last system act $a$, a 6-dimensional value-specific act vector is extracted by $f_2(v,\cdot)$. It consists of 6 indicators about $v$: (1) whether $a$ requests the slot $s$. (2) whether $a$ confirms $v$ as liked. (3) whether $a$ confirms $v$ as disliked. (4) whether $a$ confirms other values in slot $s$. (5) whether  $a$ informs $v$. (6) whether none of above holds.

3. Current user utterance $u$ is converted into a value-specific embedding matrix by $f_3(v,\cdot)$ operation. Suppose current utterance $u$ contains $k_{u}$ words $u_1,u_2,\dots,u_{k_{u}}$. The embedding operation $e(\cdot)$ converts a word into a vector in $\mathbb{R}^{d}$, where $d$ denotes the word embedding dimension. 
Let $X\in \mathbb{R}^{k_{u}\times d}$ denote the word embedding matrix of user utterance $u$. 
We convert $u$ into value-specific embedding matrix $f_3(v,u)\in \mathbb{R}^{k_{u}\times (d+2)}$ by concatenating (denoted by $\oplus$) $X$ with two other vectors as follows:
\begin{equation}
\label{value-specific matrix}
f_3(v,u) =X \oplus x_{dot}(v,u) \oplus x_{str}(v,u)
\end{equation}
where $x_{dot}(v,u) = \sigma\left(w_1 \left(X e(v)\right)+b_1\right)$. $w_1,b_1$ are scalar trainable paramenters. 
$\sigma(\cdot)$ denotes the sigmoid function. 
$x_{dot}(v,u)\in \mathbb{R}^{k_{u}}$ stores the result of dot products between the embedding of $v$ and every word embedding of $u$ followed by a nonlinear transform to $[0,1]$.
Since dot products between similar words' embeddings tend to be large, so this term reflects the extent of $v$ appearing in $u$.

$x_{str}(v,u)\in \mathbb{R}^{k_{u}}$ denotes the string-matching binary vector, in which the $i$-th element taking 1 if the word $u_i$ matches $v$ and 0 otherwise.
We use words in the ontology if there's no semantic dictionary. Semantic dictionary is hand-designed task-related synonym lists.
By using both $x_{dot}$ and $x_{str}$, we can combine both the advantages of hand-crafted semantic dictionary and pre-trained word vectors.

\begin{figure}[htp]
\centering
\centerline{\includegraphics[width=6cm]{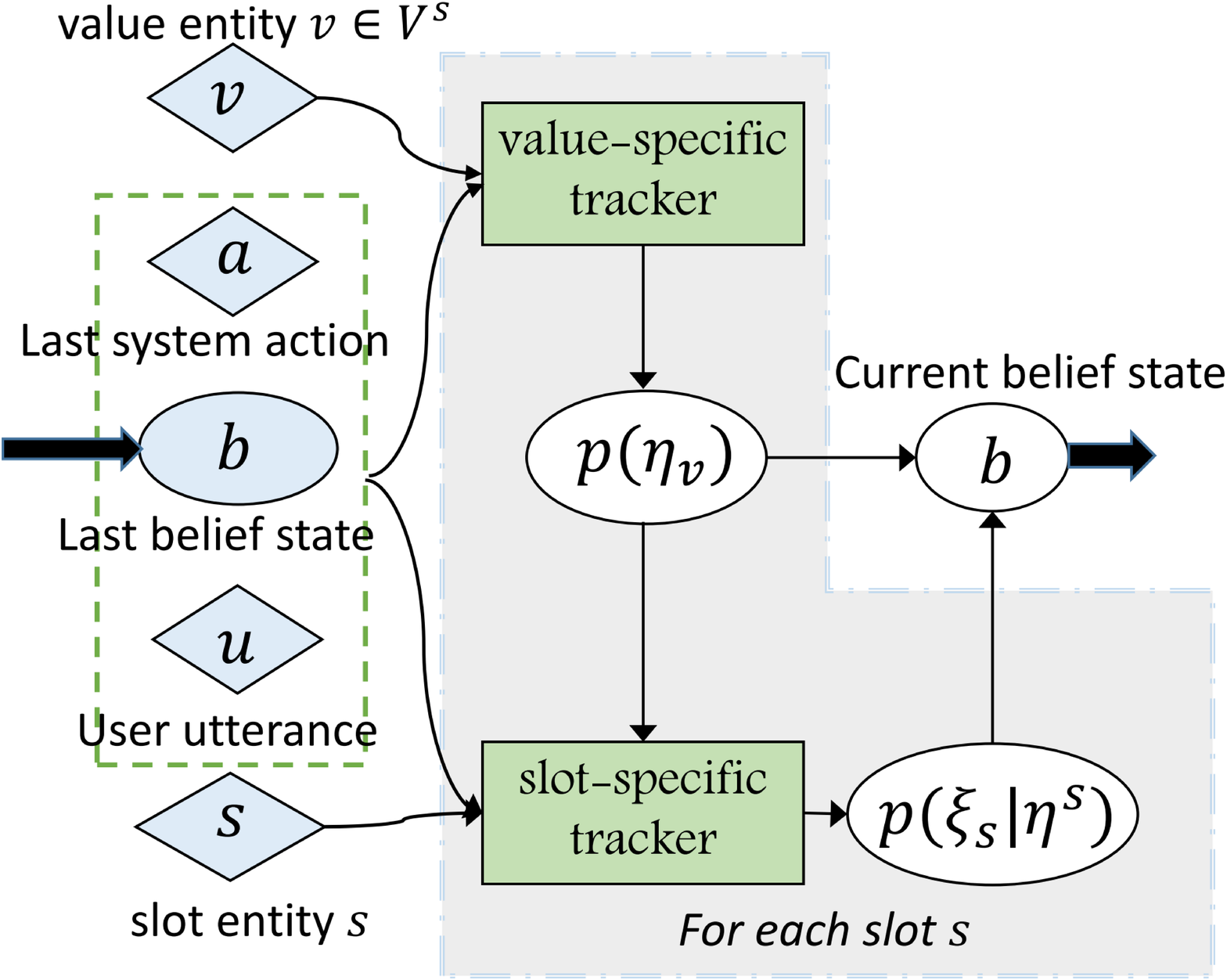}}
\caption{The architecture of EDST}
\label{fig:EDST}
\end{figure}

Second, based on $f_1(v,b) \in \mathbb{R}^{3},f_2(v,a)\in \mathbb{R}^{6},f_3(v,u)\in \mathbb{R}^{k_{u}\times (d+2)}$,
we further extract the value-specific features for belief updating of $p(\eta_v)$.
We first use a CNN module, similar to \cite{Mrk2016Neural}.
$L$ convolutional filters of window sizes $1,2,3$ are applied to value-specific embedding matrix $f_3(v,u)$. 
The convolutions are followed by the ReLU activation function and max-pooling to produce summary vectors for these $n$-gram ($n=1,2,3$) like features.
The three summary vectors are concatenated into one vector as the output of the CNN.
In the following, we use $\phi_{dim}^{sigmoid}(\cdot)$ and $\phi_{dim}^{softmax}(\cdot)$ to denote a fully-connected neural network with one hidden layer having the same size as input layer, where the output layer is of size $dim$, and is sigmoid and softmax respectively.
The value-specific features are then obtained as follows:
\begin{equation}
\begin{split}
g_f =& \phi_3^{sigmoid}\left(\text{CNN}(f_3(v,u))\right)\\
h_f=& f_1(v,b) \otimes g_f\\
r_{f,i}=& \boldsymbol{1}(f_2(v,a))[i]=1)\cdot \text{CNN}(f_3(u,v))\ i=1,2,\cdots,6
\end{split}
\end{equation}
where $\boldsymbol{1}()$ is indicator function, $f_2(v,a)[i]$ is the $i$-th element of $f_2(v,a)$, $\otimes$ denotes element-wise product. Note that here $g_f$ and $f_2(v,a)$ act like gates to control the information flow and adapt CNNs for different system acts, which is helpful for dealing with goal changes and expression variations in the conversation. 

Finally, we concatenate the above features into one vector, and feed it into a fully-connected network to update $p(\eta_v)$ by:
\begin{equation}
\phi_3^{softmax}\left(h_f \oplus r_{f,1} \oplus \dots \oplus r_{f,6}\right)
\end{equation}

\subsection{Learning and Inference}
\label{sec:model learning and inference}
SST has the same architecture as VST except that value entity $v$ is replaced by slot entity $s$. Also note that according to our EDST label settings, we have:
\begin{equation}
\begin{split}
p(\xi_{s}=\texttt{MENTIONED}|\exists v\in V^s, \eta_{v} \ne \texttt{NOT\_MEN})=1\\
p(\xi_{s}=\texttt{MENTIONED}|\forall v \in V^s,  \eta_{v}=\texttt{NOT\_MEN})=0\\
\end{split}	
\end{equation}
Therefore for updating $p(\xi_s|\eta^s)$, SST in fact only needs to infer about \texttt{DONT\_CARE, NOT\_MEN}. Details are omitted to save space.

During learning, a complete dialog data is decomposed into multiple turn data, each is then transformed into two types of data for training VST and SST respectively. 
During inference, we first use VST to obtain $p(\eta_v)$, choose the maximum a posterior (MAP) value label for each value $v\in V^s$, and then calculate $p(\xi_s|\eta^s)$ with SST to obtain the complete belief state $p(\xi_s,\eta^s)$.

We separately maintain a VST and a SST for each informable slot, and a VST for each requestable slot.
As requestable slots serve to model single-turn user
queries, slot tracking across turns and value tracking are not needed.
We remove cross-turn input $a$ and $b$, feed only current user utterace $u$ and the requestable slot label $s$ to SST, which infer whether this requestable slot is mentioned or not.

\section{EXPERIMENT}
\label{sec:experimet}
EDST is mainly designed for Iqiyi dialog dataset, but it can also be used for WOZ2.0 and DSTC2. In the latter case, we have only two value labels \texttt{MENTIONED,NOT\_MENTIONED}; during inference, we choose the MAP value label for each value $v\in V^s$, and then choose the most probable value, since a slot can take only a single value.

Glove.300d \cite{Pennington2014Glove} vectors are used for word embedding. For all experiments, CNN filters $L=50$, so the output of CNN is a vector of size 150.
Dropout with 50\% rate is used in intermediate NN layers, and gradient clipping is applied with max global norm 5 to handle exploding gradients. 
All models are trained under cross-entropy loss with Adam optimizer \cite{Kingma2014Adam}, early-stopping is employed on validation set to prevent over-fitting. 
Careful design of minibatch sampling\footnote{Minibatch size is 256 and 64 for VST and SST respectively. For both VST and SST, the ratio of positive and negative training samples is 1:7 for WOZ2.0, DSTC2 and Iqiyi, except that the ratio is 0.7:0.3:7 for Iqiyi VST of \texttt{LIKE,DISLIKE,NOT\_MENTIONED}.} 
is used to solve the label bias problem.
The  evaluation metrics are Joint Goal and Request, which is the turn-level accuracy of tracking results for all informable slots and all requestable slots respectively.

\subsection{Results on WOZ2.0 dataset}
In WOZ2.0, since turn-level labels are available, we remove last belief state $b$ from input in EDST. Note that using turn-level based labels usually gives better results than using accumulated belief state labels, since the model do not have to learn the effect of last belief state. 
A rule-based tracking scheme is used to accumulate the turn-level prediction: substitute the old value label with new one if it is not labeled as \texttt{NOT\_MEN}.
It can been seen from the results in Table \ref{WOZ} that a slight gain in turn-level goal brings a large improvement on the Joint Goal. 
Our model outperforms all other methods and achieve the best results when combining the advatages of both semantic dictionary and pre-trained word vectors.
\begin{table}[htp]
	\centering
	\begin{tabular}{|l|l|l|l|}
		\hline
		Model & Turn-level Goal & Joint Goal & Request\\
		\hline
		EDST\small+dict. & \textbf{92.8} & \textbf{87.5} & \textbf{95.3}\\
		EDST &  91.6 & 85.2 & 95.2\\
		EDST\small-spec. & 85.5 & 63.0 & 90.5\\
		\hline
		RNN \cite{Mrk2016Neural}& -- & 70.8 &87.1\\
		RNN\small+dict. \cite{Mrk2016Neural} & -- &83.7&87.6\\
		NBT \cite{Mrk2016Neural} &--&84.4&91.6\\
		\hline
	\end{tabular}
	\caption{Results for delexicalisation-based RNN, NBT and EDST on WOZ2.0. {\small+dict} means using the semantic dictionary, {\small-spec.} means we feed the original word embedding matrix $X$ instead of value-specific embedding matrix $f_3(v,u)$ to CNNs.}
	\vspace{-0.5cm}
	\label{WOZ}
\end{table}

\subsection{Results on DSTC2 dataset}
In DSTC2, we need to handle ASR problem. We train EDST on the transcripts and evaluate on the testset's ASR $N$-best lists. Let $P_i$ be the posterior probability for the $i$-th hypothesis at current turn, $\text{hyp}_i$ be the $i$-th hypothesis utterance, the belief state is represented as follows:
\begin{equation}
p(\xi, \eta|\text{ASR})=\sum_{i=1}^{N}P_i \ p(\xi, \eta|\text{hyp}_i)
\end{equation}
Because the language usage in the DSTC2 dataset is less rich, adding semantic dictionary is less useful than in WOZ2.0. Table \ref{DSTC2} shows the results of different models trained only with transcripts. Our EDST achieves superior results.
\begin{table}[htp]
	\setlength{\abovecaptionskip}{0.cm}
	\setlength{\belowcaptionskip}{-0.cm}
	\centering
	\begin{tabular}{|l|l|l|}
		\hline
		Model  & Joint Goal & Request\\
		\hline
		EDST & 73.9 & \textbf{96.6}\\
		\hline
		RNN\small+dict. \cite{Mrk2016Neural} & 72.9 & 95.7\\
		NBT \cite{Mrk2016Neural}&73.4&96.5\\
		MemN2N \cite{Perez2016Dialog} & \textbf{74}& --\\
		\hline
	\end{tabular}
	\caption{Results for NBT, MemN2N and EDST on DSTC2.}
	\label{DSTC2}
	\vspace{-0.5cm}
\end{table}

\subsection{Results on Iqiyi dataset}
For Iqiyi dialog dataset, since there is no off-the-self DST model suitable for the new dialog state labels, we construct a template-based baseline to compare with our EDST. All the template are extracted from the training and validation set\footnote{For Iqiyi dataset, the ratio of training, validation and testing size is 3:1:1.} by delexicalisation \cite{wenN2N17}, and used in testing with fuzzy string matching. We employ jieba toolbox\footnote{https://github.com/fxsjy/jieba} to segment Chinese words, and learn 25-dimensional semantically specialised word vectors using a method similar to \cite{Wieting2015From}. Final results are shown in Table \ref{Iqiyi}. We find that DST on Iqiyi dataset is more difficult mainly due to lexicon variations and task variety.
\begin{table}[htp]
	\centering
	\begin{tabular}{|l|l|l|}
		\hline
		Model  & Joint Goal & Request\\
		\hline
		EDST\small+dict.& \textbf{70.1} & \textbf{97.4}\\
		EDST & 63.6 & 97.0\\
		\hline
		template\small+dict. & 46.3 & 82.5\\
		\hline
	\end{tabular}
	\caption{Results on Iqiyi dialog dataset.}
	\label{Iqiyi}
	\vspace{-0.5cm}
\end{table}

\section{RELATED WORK}
\label{sec:related work}

Recent DST studies mainly focus on using deep neural networks. Initially, a word-based RNN with n-gram features is proposed in \cite{Henderson2014Word,Henderson2015Robust}.
It has been shown in \cite{wenN2N17} that employing CNN features with RNN can yield better results. In \cite{Mrk2016Neural}, given fine-trained semantic word vectors and turn-level labels, rule-based DST with CNNs outperforms traditional RNN models. A hybrid DST is built in \cite{Vodol2017Hybrid}, which consists of both rule-based part and machine-learning part. Novel structures such as attention-based Seq2Seq \cite{Hori2017Dialog} and Memory Network \cite{Perez2016Dialog} are also studied. 

Our EDST model is motivated by the Neural Belief Tracker (NBT) \cite{Mrk2016Neural} and delexicalisation-based RNN belief tracker \cite{wenN2N17}. NBT utilizes fine-pretrained word vectors to reduce the burden of building semantic dictionary, and delexicalisation-based RNN does not need turn-level labels. Our new EDST model combines the strengths of the two models.  
Label dependency modeling has also been studied. \cite{Xu2014Convolutional} proposes a CNN-based triangular CRF for sentence-level
optimization, \cite{liu2015recurrent} combines the Jordan-type and Elman-type RNNs
to predict the outputs depending on the last labels. In this work, to deal
with the goal change problem, we build a CNN acting as a gate on
last belief state to learn how to update the belief state.

\section{CONCLUSION AND FUTURE WORK}
\label{sec:conclusion}

In this work, we first introduce a new dialog dataset Iqiyi with enriched DST labels which can represent polarity preference and multi-values, then we develop a novel RNN-based EDST and achieve superior performances on WOZ2.0, DSTC2 and our Iqiyi dataset.

There are interesting future works so as to achieve more flexible conversational information access: building the whole dialog system based on EDST; unifying the searching and enquiring tasks in a better way; incorporating semantic parsing.

\bibliographystyle{IEEEbib}
\bibliography{refs}

\end{document}